# SENTIMENT FRAMES FOR ATTITUDE EXTRACTION IN RUSSIAN

**Loukachevitch N. V.** (louk_nat@mail.ru)

Lomonosov Moscow State University, Moscow, Russia

**Rusnachenko N. L.** (kolyarus@yandex.ru)

Bauman Moscow State Technical University, Moscow, Russia

Texts can convey several types of inter-related information concerning opinions and attitudes. Such information includes the author's attitude towards mentioned entities, attitudes of the entities towards each other, positive and negative effects on the entities in the described situations. In this paper, we described the lexicon RuSentiFrames for Russian, where predicate words and expressions are collected and linked to so-called sentiment frames conveying several types of presupposed information on attitudes and effects. We applied the created frames in the task of extracting attitudes from a large news collection.

# ОЦЕНОЧНЫЕ ФРЕЙМЫ ДЛЯ ИЗВЛЕЧЕНИЯ ОЦЕНОЧНЫХ ОТНОШЕНИЙ ИЗ ТЕКСТОВ НА РУССКОМ ЯЗЫКЕ

**Лукашевич Н. В.** (louk_nat@mail.ru)

МГУ имени М. В. Ломоносова, Москва, Россия

**Русначенко Н. Л.** (kolyarus@yandex.ru)

МГТУ имени Н. Э. Баумана, Москва, Россия

## 1. Introduction

Sentiment analysis of texts is one of the most important directions in natural language processing research. Numerous papers were devoted to problems of automatic classification of whole texts or text fragments to two, three or five classes according to the opinion found in this text. Currently, researchers began to consider several types of interrelated phenomena that previously had considered as the same type of sentiment attitudes.

[Mohammad 2016] indicates the following different subtypes of phenomena interplaying with author sentiment: speaker emotion state vs. polarity of opinion mentioned in a sentence; success or failure vs. opinion about the success of failure; neutral reporting of a negative (war) or positive (celebration) event etc. Additionally, an author can mention some negative or positive relations between entities without conveying his/her own stance [Loukachevitch, Rusnachenko, 2017]. [Feng et al. 2013] consider polar connotations (sentiment associations) of objective words such as *unemployment*, *war* (negative connotations) or *human rights, sun* (positive connotations). These words do not express direct sentiments but may accompany or contradict attitudes conveyed in the text. [Choi et al. 2014] consider the interaction between described good (bad) effects and conveyed polarities.

Thus, we can see a scope of various phenomena related to sentiment analysis, but most existing sentiment vocabularies have a simple structure as lists of words or expressions with positive or negative sentiment scores. For example, in the news title "White House Blocked 2018 Statement Condemning Russia", both named entities 'White House' and 'Russia' are located in the context of negative words *block* and *condemn* (according to the MPQA lexicon [Wilson et al., 2005]) but we do not infer any author's stance to White House or Russia. Besides, we can suppose some positive effects from the described situation to Russia and no sentiments or effects on White House.

At the same time, similar language means can be used to convey implicit attitudes of the author [Liu, 2012]; [Nozza et al., 2017]. For example, in the sentence "He approved the bombing of civilians" the author of the text is negative towards the subject. Here the following issues interplay: a positive attitude to civilians presupposed for most people, a negative action towards civilians and approving this event by the subject. The interaction of the above-mentioned phenomena usually correlates with specific sentiment predicates [Rashkin et al., 2016]; [Klenner et al., 2017], which can have some presupposed information about attitudes between participants of the situation described by the predicate or effects on the participants as a result of the situation.

In this paper we consider a specific lexicon called RuSentiFrames for Russian, where predicate words and expressions are collected and linked to so-called sentiment frames conveying several types of presupposed information on attitudes and effects. We give the detailed description of RuSentiFrames and describe methods for evaluation of the created resource.

The structure of the paper is as follows. **Section 2** describes related work. In **Section 3** we present the structure of the RuSentiFrames lexicon, the principles of describing its lexical entries. **Section 4** is devoted to evaluation of the RuSentiFrames lexicon.

## 2. Related work

Most sentiment vocabularies are presented as lists of words and expressions with scores of their sentiment [Wilson et al. 2005]. Some vocabularies provide also additional characteristics of the word sentiment called as 'strength'. Also sentiment scores can be assigned to specific senses of ambiguous words [Baccianella et al., 2010]; [Loukachevitch and Levchik, 2016]. For more accurate extraction of sentiment attitudes cited or expressed in texts, it is not enough to have a simple sentiment list with sentiment scores assigned to words and expressions [Neviarouskaya, 2009]; [Deng and Wiebe, 2014].

In [Rashkin et al., 2016] the authors stressed that it is important to extract implied sentiments and proposed the approach to description of so-called connotation frames for transitive verbs. The description includes three participants' roles: agent, theme, and writer. The frame includes the attitude polarity of participants to each other (positive, negative, or neutral), the effect of the situation to agent or theme, mental states and values of the participants. For experiments, 1,000 most frequent English verbs were extracted from a corpus. These verbs were provided with five example sentences constructed from most frequently seen Subject-Verb-Object triples in Google syntactic ngrams and were annotated by crowdsourcers. The obtained values were averaged.

[Klenner et al. 2014], [2016] described a verb resource for German containing verb polarity frames. Each frame consists of the subcategorization frame and the polarity (positive or negative) effects associated with the roles. Also so-called verb signature is assigned. The verb signature indicates the factuality of roles in dependence on various factors (such as negation, mood, etc.). By 2017, 1,500 verb-polarity frames for 1,100 verbs were described. Some nominalizations (for example, *destruction*) have been also considered as frame entries [Klenner et al., 2017].

[Deng et al. 2014] considered events that positively or negatively affect entities (goodFor/badFor). For example, lowering something is bad for this something, but creating something is good for this something. This paper also describes the sentiment inference when the sentiment is conveyed towards a bad or good event.

Several Russian sentiment lexicons of sentiment words with scores have been published. In [Chetviorkin and Loukachevitch, 2012], automatically generated Russian sentiment lexicon in the domain of products and services (ProductSentiRus) is described. The ProductSentiRus is obtained by application of a supervised model to user's review collections in several domains. It is presented as a list of 5,000 words ordered by the decreased probability of their sentiment orientation without any positive or negative labels.

The general Russian lexicon of sentiment words and expressions, RuSentiLex, was created in a semi-automatic way [Loukachevitch and Levchik, 2016]. In structure, RuSentiLex is also a list of words and expressions having several attributes. The entries of the RuSentiLex lexicon are classified according to four sentiment categories (positive, negative, neutral, or positive/negative) and three sources of sentiment (opinion, emotion, or fact). The words in the lexicon having different sentiment orientations in different senses are linked to appropriate concepts of the Russian thesaurus RuThes [Loukachevitch and Dobrov, 2014], which can help disambiguate sentiment ambiguity in specific domains or contexts.

Russian Sentiment Lexicon Linis Crowd has been created via crowdsourcing [Koltsova et al. 2016]. The lexicon is aimed at detecting sentiment in user-generated content (blogs, social media) related to social and political issues. Each word was assessed by at least three volunteers in the context of three different texts and scored from −2 (negative) to +2 (positive).

Several international lexicons were automatically constructed for Russian. The Chen-Skiena lexicon (2,876 words) [Chen and Skiena, 2014] was generated for 136 languages via graph propagation from seed words, including Russian. [Mohammad and Turney 2013] generated the Russian variant of the EmoLex lexicon with automatic translation from the English lexicon obtained by crowdsourcing (4,412 Russian words). [Kotelnikov et al. 2018] studied available Russian sentiment lexicons and found that all the lexicons have relatively small intersection with each other. They compared the above-mentioned Russian lexicons as features in machine-learning text categorization of user's reviews in several domains using the SVM method. It was found that the best results of classification using a single lexicon in all domains were obtained with ProductSentiRus [Chetviorkin and Loukachevitch 2012]. The union of all lexicons gives slightly better results.

## 3. Sentiment Frames for Sentiment Analysis in Russian

Russian Sentiment Lexicon RuSentiFrames describes sentiments and connotations conveyed with a predicate word in either verbal or nominal form[1].

### 3.1. General Structure of Sentiment Frames

In this study, *sentiment frame* is a set of positive or negative associations (connotations) related to a predicate word or expression. A predicate usually describes a situation with some participants. The types of connotations that are conveyed in sentiment frames are as follows:
- attitude of the author of the text towards mentioned participants,
- positive or negative sentiment between participants,
- positive or negative effects on participants,
- positive or negative mental states of participants related to the described situation.

To describe participants of a situation, we should designate predicate-specific roles. There are several approaches to sets of semantic roles such as universal sets [Jackendoff, 1992], frame elements as specific roles for each frame as in FrameNet [Fillmore and Baker, 2001], or enumerated roles as in PropBank [Palmer et al., 2005] and AMR representation [Banarescu et al., 2013]. It is very difficult to choose an appropriate universal set of semantic roles. Therefore we accepted the approach of PropBank. In this approach, individual verb's semantic arguments are numbered, beginning with zero. For a particular verb, Arg0 is generally the argument exhibiting features of a Prototypical Agent [Dowty, 1991], while Arg1 is a Prototypical Patient or Theme.

All assertions are provided with the score of confidence, which currently has two values: 1, if we believe that this assertion is true almost always, or 0.7, if we consider the assertion as default. We do not describe assertions about neutral sentiment, effect and state of participants.

**Figure 1** presents the attitudes and effects for the *condemn* frame:

---

[1]   https://github.com/nicolay-r/RuSentiFrames/tree/v2.0

```
frame: осудить (to condemn)
    "roles": {
                "A0": "who condemns",
                "A1": "who is condemned",
                "A2": "grounds for condemnation",
                "A3": "punishment"},
    "polarity":
                [["A0"," A1","neg",1.0],
                 ["A0"," A2","neg",1.0],
                 ["A0"," A3","neg",1.0],
                 ["A1"," A0","neg",1.0],
                 ["A1"," A3","neg",1.0]],
    "effect": [["A1","-",1.0]],
    "state":  [["A1","neg",1.0]]}
```

**Fig. 1:** Frame example

This frame means that a condemner A0 is negative to a condemned person A1, to the grounds of condemnation A2 and knows that the punishment A3 is also something negative. A1 is negative to A0 (for the condemnation) and to punishment. The effect on A1 and the state of A1 are negative. We denote effects with signs "−" or "+" to highlight that this information is different from sentiment attitudes. For the example verb, we cannot guess the author's opinions on the described situation and their participants therefore the corresponding assertions are absent in the frame.

If words or expressions are semantically related and have the same roles and associated connotations, we assign them to the same frame. Currently, we do not associate roles with syntactic means of their expression planning to gather this information in a semi-automatic manner.

### 3.2. Procedure of Frame Description

Experts describe frames using the following steps:

- choosing a target word with positive or negative connotations,
- search for semantically related words (synonyms, hyponyms) and expressions with similar connotations,
- introducing the main roles of a situation described with the target word,
- description of connotations of relations between participants, an author to participants, effects and states. The expert should check the usage of the target words analyzing sentence examples in contemporary Russian texts.

In complicated cases, the following additional guidelines are applied.

In description of connotations, social and human rights values should be accounted for, if relevant. For instance, this means that the prior author's attitude toward the agent (A0) in the killing event is negative. In examples such as "The police killed the terrorist", possible positive polarity of the sentence can be felt. But we suppose that such a polarity arises from the negative word "terrorist" and a common-sense rule that if a negative action is directed to a negative entity then the whole situation is considered as positive [Deng and Wiebe, 2014]. In any case, in the above-mentioned example we do not deal with the prior connotations but with their transformations in the context of negative participants. Such rules are planned to add to be utilized together with the RuSentiFrames lexicon.

Sometimes it is very useful to search for examples of usages of a target word with positive and negative participants and check how polarities change. For example, in Russian there is verb *укокошить* (finish off as to kill), which means a colloquial expression for killing with possible negative polarity to the object of the situation (A1). We can try to search for or construct sentences with positive words (such as *герой* (*hero*)) as the A1 role: "*укокошить героя*" (? *to finish the hero*). It is interesting that in such a context for Russian no negativity of the situation is conveyed. The resulted polarity looks like a negative irony towards to *hero*, the positiveness of *hero* is decreased. Thus, we can conclude that the connotations with the A1 role of *укокошить* are really negative.

### 3.3. RuSentiFrames Entries: Single Words and Expressions

We formed the list of words for the description in RuSentiFrames in the following way.

At first, we included predicate words and expressions from the RuSentiLex lexicon [Loukachevitch and Levchik, 2016], Besides, we looked through the word frequency list obtained from a news text collection to find words conveying good or bad effects: *создать* (*to create*), *понизить* (*to lower*), *рост* (*growth*), *увеличивать* (*to enhance*), etc., because these words are usually not included into sentiment lists. At last, we analyzed news reports about war actions collecting words and expressions with positive and negative effects.

The created frames are associated not only with a single entry but with a "family" of related words and expressions, which have the same attitudes. As a result, the following lexical units can be associated with a sentiment frame:

- single words: mainly verbs and nouns,
- idioms: *вешать лапшу на уши* (to hang noodles on the ears—to lie), *взять за горло* (to take by the throat—to press);
- light verb constructions: *нанести вред* (inflict harm), *нанести обиду* (cause offence), *нанести поражение* (inflict a defeat), etc.;
- verbs (or nouns) with prepositions. Sentiment attitudes conveyed by some words can significantly depend on a preposition used with this word: for example, *выступать против* (to speak against—to oppose), *завязывать с* (to stop doing something);
- compositional expressions synonymous to a lexical unit or multiword expression, which has the sentiment frame.

For example, the sentiment frame called *Запретить* (to forbid) contains 53 text entries including such expressions as *налагать запрет* (to impose a ban), *наложение запрета* (imposition of a ban), *закрывать доступ* (to close access), *закрытие доступа* (closing access), *прекращение доступа* (*termination* of access), *прекратить доступ* (to terminate access), *налагать вето* (to impose veto), etc.

Some frame entries are ambiguous. In such cases, they are attached to different frames, for example, Russian verb *выгореть* can mean 'to have success' and 'to be destroyed by fire'. Currently, it is difficult to automatically disambiguate such words but they should be assigned to relevant frames for future studies.

Currently, RuSentiFrames contains 311 frames with more than 7K associated frame entries. **Table 1** shows the distribution of the RuSentiFrames entries according to parts of speech and other characteristics.

**Table 1:** Quantitative characteristics of the RuSentiFrames entries

| Type of lexical unit | Number |
|---|---:|
| Verbs | 3,239 |
| Nouns | 986 |
| Phrases | 2,551 |
| Other | 12 |
| Unique entries | 6,788 |
| Total entries | 7,034 |

### 3.4. Classes of Frames

Most frames contain several types of slots, including the attitudes between participants, the authors' attitude, and effects on the participants. Some frames convey only information about relations between participants, but the information about the author's attitudes towards the participants and the event is absent. For example, for verbs *надеяться* (to rely on) or *создать* (to create), the first participant (A0) is positive towards the second participant (A1).

Some words are very similar in sense but differ in the author's attitude to a participant. For example, Russian verbs *наказать* and *карать* are very similar in sense (to punish). But for verb *наказывать*, the author seems neutral to the situation in default, but for *карать*, the author seems positive to the agent of the situation (A0) and negative to A1 (who is punished).

**Table 2** describes the distribution of frame entries according to sentiments between main participants of the situation and from the author to the participants. **Table 3** describes the distribution of the RuSentiFrames entries according to effects on main participants A0 and A1. Some frames contain up to 4 roles associated with some attitudes (from the author or from the other participants) or effects, for example, frames *осудить* (to condemn), *выиграть* (to win), *выгнать* (to kick out), etc.

**Table 2:** The distribution of RuSentiFrames text entries according to attitudes

| Relations | Sentiment | Number |
|---|---|---|
| A0 to A1 | Pos | 2,558 |
| A0 to A1 | Neg | 3,289 |
| Author to A0 | Pos | 170 |
| Author to A0 | Neg | 1,581 |
| Author to A1 | Pos | 92 |
| Author to A1 | Neg | 249 |

**Table 3:** The distribution of RuSentiFrames text entries according to effects on main participants

| Effect | Sentiment | Number |
|---|---|---|
| A0 | Pos | 1,008 |
| A0 | Neg | 733 |
| A1 | Pos | 2,355 |
| A1 | Neg | 3,504 |

If to compare with other sentiment-oriented resources we can say that RuSentiFrames is the only structured resource for Russian, all other existing sentiment lexicons are lists of words with attributes. If to compare with connotation frames [Rashkin et al., 2016], they took only 1,000 most frequent transitive verbs. But we created a much more structured resource: more than 6 thousand words and expressions, up to four roles, the grouping of words and expressions with similar connotations to the same frames, more elaborate analysis of complex cases.

## 4. Coverage and Evaluation of RuSentiFrames

To check the agreement in the description of the created frames, two experiments were carried out. For each experiment, 200 different words (100 for each experiment) were randomly selected.

**In the first experiment,** two experts described frames for selected words in parallel using their intuition and text examples,

**In the second experiment**, one expert created frames and gave only roles (without connotations) to an annotator. The annotator gathered 10 random non-duplicate sentences for each word from different topics of the current news flow. The task of the annotator was to assign positive or negative scores to each role of the word mentioned in a sentence under analysis. The obtained scores were averaged. The average scores and connotations were compared with the original frame of the word.

We cannot estimate inter-annotator agreement using usual techniques [Bobichev and Sokolova, 2017], because any expert can add or miss some connotations, or do not reveal them in texts. We can consider two sets of connotations for each of the experts ($E_1$ and $E_2$). We calculate the intersection between two these sets $E_1 \cap E_2$, which include the same dimensions with the same positive or negative scores. Then we consider ratios:

$$R_1 = \frac{E_1 \cap E_2}{E_1} \qquad R_2 = \frac{E_1 \cap E_2}{E_2}$$

**Table 4.** Agreement between different experts in creating and annotating frames

| Measure | Expert1 vs. Expert2 | Expert to Annotator |
|---|---|---|
| $R_1$ | 0.81 | 0.82 |
| $R_2$ | 0.72 | 0.75 |
| Harmonic mean | 0.76 | 0.78 |

At last, we calculate the harmonic mean (HM) between two ratios to average them (like for F-measure). The results presented in Table 4 show that there is a considerable share of core connotations, for which opinions of experts (annotators) coincide.

To evaluate RuSentiFrames in an analytical task, we extracted frequent attitudes between named entities in the news flow and after that we could estimate if the extracted attitudes correlate with factual attitudes. We exploited a news corpus of 2017 consisting of 2.5 M news articles. Named entities were extracted with DeepPavlov library[2]. We considered sentences where at least two named entities and at least a single frame entry (internal frame entry) between them were mentioned. The corresponding frames should have positive or negative labels for A0 to A1 attitudes.

We assigned the positive sentiment score when all the polarities of the internal frame entries had the positive sentiment. Otherwise, the negative sentiment score was assigned. We also consider the frame entry polarity as inverted, when it was used with negation. **Table 5** presents the most negative attitudes found in the corpus. **Table 6** shows the most positive attitudes from the same corpus.

**Table 5.** The most negative attitudes found in the 2017 news corpus

| A0 | A1 | Frequency of co-occurrence in frames | Positive | Negative |
|---|---|---|---|---|
| Нагорный Карабах (Nagorno-Karabakh) | Азербайджан (Azerbaijan) | 123 | 0 (0%) | 123 (100%) |
| Израиль (Israel) | Дамаск (Damascus) | 41 | 0 (0%) | 41 (100%) |
| Пентагон (Pentagon) | Аль-Каеда (Al Qaeda) | 36 | 0 (0%) | 36 (100%) |
| Россия (Russia) | ИГИЛ (ISIL) | 245 | 19 (7.8%) | 226 (92.2%) |
| Киев (Kiev) | Госдума (State Duma) | 158 | 3 (1.9%) | 155 (98.1%) |
| Турция (Turkey) | ИГИЛ (ISIL) | 142 | 6 (4.2%) | 138 (95.8%) |
| Азербайджан (Azerbaijan) | Армения | 96 | 4 (4.2%) | 92 (92.8%) |

---

[2] http://docs.deeppavlov.ai/en/master/features/models/ner.html

According to the frame-based analysis, the USA-Russia relations were mainly indicated as negative (65%), the USA-Ukraine relations are mainly positive (64%), the EU-Russia relations are mainly negative (78%), the Great Britain-Russia relationships are mainly negative (59.2%). Thus we can see that the extracted attitudes correspond to factual attitudes between entities. In future, the extracted attitudes will be analyzed in a more detailed way to improve frame descriptions.

**Table 6.** The most positive attitudes found in the 2017 news corpus

| A0 | A1 | Frequency of co-occurrence in frames | Positive | Negative |
|---|---|---|---|---|
| Шойгу (Shoigu) | Путин (Putin) | 44 | 44 (100%) | 0 (0%) |
| НАТО (NATO) | Эстония (Estonia) | 30 | 30 (100%) | 0 (0%) |
| Украина (Ukraine) | МВФ (IMF) | 185 | 172 (93%) | 13 (7%) |
| Порошенко (Poroshenko) | НАТО (NATO) | 169 | 165 (97.6%) | 4 (2.4%) |
| Канада (Canada) | Украина (Ukraine) | 141 | 135 (95.7%) | 6 (4.3%) |
| Украина (Ukraine) | НАТО (NATO) | 103 | 96 (93.2%) | 7 (6.8%) |
| США (USA) | Черногория (Montenegro) | 100 | 99 (99%) | 1 (1%) |

The created lexicon has been also already used for creating a training collection in the distant supervision framework [Rusnachenko et al., 2019].

## 5. Conclusion

Texts can convey several types of inter-related information concerning opinions and attitudes. Such information can include the author's attitude towards mentioned entities, attitudes of the entities towards each other, positive and negative effects on the entities in the described situations. The effect extraction is often important for the attitude analysis because the positive attitude towards negative situation for an entity is usually means the negative attitude towards the entity, and vice versa.

In this paper, we described the lexicon RuSentiFrames for Russian, where predicate words and expressions are collected and linked to so-called sentiment frames conveying several types of presupposed information on attitudes and effects. We applied the created frames in the task of extracting attitudes from a large news collection.

## 6. Acknowledgements

The reported study is partially funded by RFBR, research project № 20-07-01059.

## References

1. *Baccianella S., Esuli A., Sebastiani F.* (2010), Sentiwordnet 3.0: an enhanced lexical resource for sentiment analysis and opinion mining. LREC-2010.—V. 10, pp. 2200–2204.
2. *Baker, C., Fillmore Ch., Lowe J.* (1998), The Berkeley FrameNetproject. In- Proceedings of COLING/ACL,,pp. 86–90, Montreal.
3. *Banarescu, L., Bonial, C., Cai, S., Georgescu, M., Griffitt, K., Hermjakob, U., Schneider, N.* (2013), Abstract meaning representation for sembanking. In Proceedings of the 7th Linguistic Annotation Workshop and Interoperability with Discourse, pp. 178–186.
4. *Bobicev, V.,Sokolova, M.* (2017), Inter-Annotator Agreement in Sentiment Analysis: Machine Learning Perspective. In Proceedings of RANLP-2017 conference, pp. 97–102.
5. *Chetviorkin, I., Loukachevitch, N.* (2012), Extraction of Russian Sentiment Lexicon for Product Meta-Domain. In COLING-2012, pp. 593–610.
6. *Chen Y., Skiena S.* (2014), Building Sentiment Lexicons for All Major Languages. In: Proceedings of the 52nd Annual Meeting of the Association for Computational Linguistics, pp. 383–389.
7. *Choi, Y., Deng, L., Wiebe, J.* (2014), Lexical acquisition for opinion inference: A sense-level lexicon of benefactive and malefactive events. In Proc. of the 5th Workshop on Computational Approaches to Subjectivity, Sentiment and Social Media Analysis, Baltimore, Maryland.Association for Computational Linguistics.
8. *Choi E., Rashkin H., Zettlemoyer L., Choi Y.* (2016), Document-level Sentiment Inference with Social, Faction, and Discourse Context. Proceedings of the 54th Annual Meeting of the Association for Computational Linguistics, ACL-2016, pp. 333–343.
9. *Deng L., Wiebe J.* (2014), Sentiment propagation via implicature constraints. Meeting of the European Chapter of the Association for Computational Linguistics (EACL-2014).
10. *Dowty D.* (1991), Thematic proto-roles and argument selection. Language, 67(3): 547–619.
11. *Feng S., Kang J. S., Kuznetsova P., Choi Y.* (2013), Connotation Lexicon: A Dash of Sentiment Beneath the Surface Meaning. In ACL (1), pp. 1774–1784.
12. *Fillmore Ch., Baker, C.* (2001), Frame semantics for text understanding. In Proceedings of NAACL WordNet and Other Lexical Resources Workshop.
13. *Jackendoff R.* (1992), Semantic structures. MIT press, 1992.
14. *Klenner M., Amsler M., Hollenstein N.* (2014), Verb polarity frames: a new resource and its application in target-specific polarity classification. In G. Faaß (Ed.), KONVENS, pp. 106–115.

15. *Klenner M, Amsler M.,* (2016), Sentiframes: A resource for verb-centered German sentiment inference. Proceedings of the Tenth International Conference on Language Resources and Evaluation (LREC 2016), European Language Resources Association (ELRA), pp. 2888–2891.
16. *Klenner M., Tuggener D., Clematide S.* (2017), Stance detection in Facebook posts of a German right-wing party. Proceedings of the 2nd Workshop on Linking Models of Lexical, Sentential and Discourse-level Semantics, pp. 31–40.
17. *Koltsova O. Y., Alexeeva S. V., Kolcov S. N.* (2016), An Opinion Word Lexicon and a Training Dataset for Russian Sentiment Analysis of Social Media. Computational Linguistics and Intellectual Technologies, pp. 277–287.
18. *Kotelnikov E., Peskisheva T., Kotelnikova A., Razova E.* (2018), Comparative Study of Publicly Available Russian Sentiment Lexicons. In Conference on Artificial Intelligence and Natural Language, pp. 139–151.
19. *Liu B.* (2012), Sentiment Analysis and Opinion Mining, Morgan and Claypool Publishers.
20. *Loukachevitch N., Dobrov B.* (2014), RuThes linguistic ontology vs. Russian wordnets. In Proceedings of the Seventh Global Wordnet Conference, pp. 154–162.
21. *Loukachevitch N., Levchik A.* (2016), Creating a General Russian Sentiment Lexicon. In Proceedings of Language Resources and Evaluation Conference LREC-2016, 2016.
22. *Loukachevitch N., Rusnachenko N.* (2018), Extracting sentiment attitudes from analytical texts. arXiv preprint arXiv:1808.08932.
23. *Mohammad S. M.* (2016), A practical guide to sentiment annotation: Challenges and solutions. Proceedings of NAACL-HLT, pp. 174–179.
24. *Mohammad S. M., Turney D. P.* (2013), Crowdsourcing a word-emotion association lexicon. Computational Intelligence 29(3): 436–465.
25. *Neviarouskaya A., Prendinger H., Ishizuka M.* (2009), Semantically distinct verb classes involved in sentiment analysis. IADIS AC (1), pp. 27–35.
26. *Nozza D., Fersini E., Messina E.* (2017), A Multi-View Sentiment Corpus. In Proceedings of EACL-2017, pp. 273–280.
27. *Palmer M., Gildea D., Kingsbury P.* (2005), The proposition bank: An annotated corpus of semantic roles. Computational linguistics: 71–106.
28. *Rashkin H., Singh S., Choi Y.* (2016), Connotation Frames: A Data driven Investigation. Proceedings of Association for Computational Linguistics Conference ACL-2016, pp. 311–322.
29. *Rusnachenko N., Loukachevitch, N., Tutubalina, E.* (2019), Distant supervision for sentiment attitude extraction. In Proceedings of the International Conference on Recent Advances in Natural Language Processing (RANLP 2019), pp. 1022–1030.
30. *Wilson T., Wiebe J., Hoffmann P.* (2005), Recognizing contextual polarity in phrase-level sentiment analysis. Proceedings of the conference on human language technology and empirical methods in natural language processing, pp. 347–354.